\documentclass{article}
\usepackage{spconf,amsmath,epsfig}
\usepackage{multirow} 
\usepackage{amsfonts}
\usepackage{booktabs}
\usepackage{threeparttable}
\usepackage{bm}
\usepackage[marginal]{footmisc}

\begin{document}\sloppy

\def\x{{\bm x}}
\def\L{{\cal L}}

\title{MatchingGAN: Matching-based Few-shot Image Generation}
%
\name{Yan Hong, Li Niu, Jianfu Zhang, Liqing Zhang}
\address{}

\maketitle


\begin{abstract}
To generate new images for a given category, most deep generative models require abundant training images from this category, which are often too expensive to acquire. To achieve the goal of generation based on only a few images, we propose matching-based Generative Adversarial Network (GAN) for few-shot generation, which includes a matching generator and a matching discriminator. Matching generator can match random vectors with a few conditional images from the same category and generate new images for this category based on the fused features. The matching discriminator extends conventional GAN discriminator by matching the feature of generated image with the fused feature of conditional images. Extensive experiments on three datasets demonstrate the effectiveness of our proposed method 
\end{abstract}

\section{Introduction}
\label{sec:intro}
Deep generative models like Variational Auto-Encoder (VAE)~\cite{bao2017cvae} and Generative Adversarial Network (GAN)~\cite{goodfellow2014generative,arjovsky2017wasserstein} have made tremendous advance in generation problems. However, to generate new samples for a given category, the training of deep generative models relies on abundant labeled training data from this category. To achieve the goal of generation based on a few images, several few-shot generation methods~\cite{clouatre2019figr,bartunov2018few} have been proposed, which aim to generate more images from certain category based on a few images from this category. The model is trained on seen categories in the training stage, and then applied to generate more images for unseen categories which do not appear in the training stage. In this way, meta information or metric information could be transferred from seen categories to unseen categories, which makes few-shot generation possible for unseen categories. Existing few-shot generation methods can be categorized into feature generation methods~\cite{schwartz2018delta} and image generation methods~\cite{antoniou2017data}. Specifically, given a few images (\emph{resp.}, features) from one category, few-shot image (\emph{resp.}, feature) generation methods target at generating more images (\emph{resp.}, features) from this category. 

In this paper, we focus on few-shot image generation, which is much more challenging than few-shot feature generation. Generated images can augment training data and facilitate downstream tasks like few-shot classification. To the best of our knowledge, there are quite limited works~\cite{clouatre2019figr,bartunov2018few,antoniou2017data} on few-shot image generation. For example, FIGR~\cite{clouatre2019figr} incorporates meta-learning~\cite{nichol2018first} into GAN to learn the data distribution of a few images from the same category. However, the quality of generated images is inferior and the model needs to be fine-tuned on the images from unseen categories at the test phase. GMN~\cite{bartunov2018few} combines matching procedure and VAE to generate novel images conditioned on a few images. However, due to the weakness of VAE, the images generated by GMN are vague and unrealistic. DAGAN~\cite{antoniou2017data} was designed to generate diverse images based on a single conditional image by injecting random noise into decoder. However, the diversity of generated images brought by injected noise is quite limited. Besides, DAGAN is conditioned on a single image, and thus fails to exploit the relationship among multiple images from the same category.

Considering the drawbacks of the above few-shot image generation methods, we propose 
Matching-based Generative Adversarial Network (MatchingGAN), inspired by matching-based methods~\cite{vinyals2016matching,bartunov2018few} which prove that matching procedure learned from training samples of seen categories can be adapted to unseen categories. Our MatchingGAN can fully exploit multiple conditional images from the same category to generate more diverse and realistic images by virtue of the combination of adversarial learning and matching procedure. In detail, our MatchingGAN is comprised of a matching generator and a matching discriminator. In the matching generator, we project a random vector and a few conditional images from one seen category into a common matching space, and calculate the similarity scores, which are used as interpolation coefficients to fuse features of conditional images to generate new images belonging to this seen category. Through the matching procedure, we can learn reasonable interpolation coefficients which determine how much information we should borrow from each conditional image. In the matching discriminator, we not only distinguish real images from fake images as done by conventional GAN discriminator, but also match the discriminative feature of generated image with the fused discriminative feature of conditional images to ensure that the generated image contains the interpolated information of conditional images. In the test phase, given a few conditional images from one unseen category,  we can feed sampled random vectors into matching generator to generate numerous diverse and realistic images for this unseen category.

Our major contributions can be summarized as follows: 1) We propose a novel few-shot image generation method by combining matching procedure and adversarial learning; 2) Technically, we design a matching generator and a matching discriminator for our MatchingGAN; 3) Comprehensive generation and classification experiments on three datasets demonstrate the effectiveness of our MatchingGAN.

\section{Related Work}
\label{sec:related}

\textbf{Generative adversarial network}
Generative Adversarial Network (GAN)~\cite{goodfellow2014generative} was proposed to discriminate real samples from fake samples and generate more realistic samples. In the early stage, unconditional GANs~\cite{arjovsky2017wasserstein,miyato2018spectral} use random vectors to generate realistic samples based on the learned distribution of training samples. Then, GANs conditioned on a single image~\cite{miyato2018cgans,lu2018image,antoniou2017data} were proposed to transform the conditional image to a target image by using adversarial learning. Recently, a few conditional GANs attempted to leverage more than one image to accomplish more challenging tasks, such as few-shot image translation~\cite{liu2019few} and few-shot image generation~\cite{clouatre2019figr}. In this paper, we choose to combine matching procedure and GAN to capture the variance of a few conditional images to generate diverse images. 

\noindent\textbf{Few-shot learning}
Existing few-shot learning methods are mainly classified into three categories: metric-based methods~\cite{vinyals2016matching,sung2018learning,zheng2019relation}, model-based methods~\cite{santoro2016one}, and optimization-based methods~\cite{finn2017model,nichol2018first,du2019low}. Our work is most related to metric-based methods~\cite{vinyals2016matching,sung2018learning,zheng2019relation}, which can learn well-tailored metric to classify samples of unseen categories by comparing with a few labeled images from unseen categories. The process of learning metric well-tailored for specific tasks is defined as matching procedure.

Our MatchingGAN employs matching procedure to learn reasonable interpolation coefficients to fuse the features of conditional images.

\noindent\textbf{Data augmentation}
Data augmentation~\cite{Krizhevsky2012ImageNet} aims to generate more samples based on the given samples. Early data augmentation tricks such as shifts, rotations or shears can only produce limited diversity. In contrast, deep generative models could generate more diverse samples for data augmentation, including feature augmentation~\cite{hariharan2017low,wang2018low,schwartz2018delta} and image augmentation~\cite{antoniou2017data}. Our method can be deemed as an image augmentation method. For image augmentation, previous few-shot image generation methods like FIGR\cite{clouatre2019figr}, GMN~\cite{bartunov2018few}, DAGAN~\cite{antoniou2017data} attempted to generate images based on a single or a few conditional images.

Besides, few-shot image translation method like FUNIT~\cite{liu2019few} intended to translate images from seen categories to unseen categories. However, in the testing phase, few-shot image translation relies on the images from seen categories to generate new images for unseen categories, which is a different task from few-shot generation. In this work, we propose a novel few-shot image generation method which could be used for data augmentation.

\begin{figure*}
\begin{center}
\includegraphics[scale=0.49]{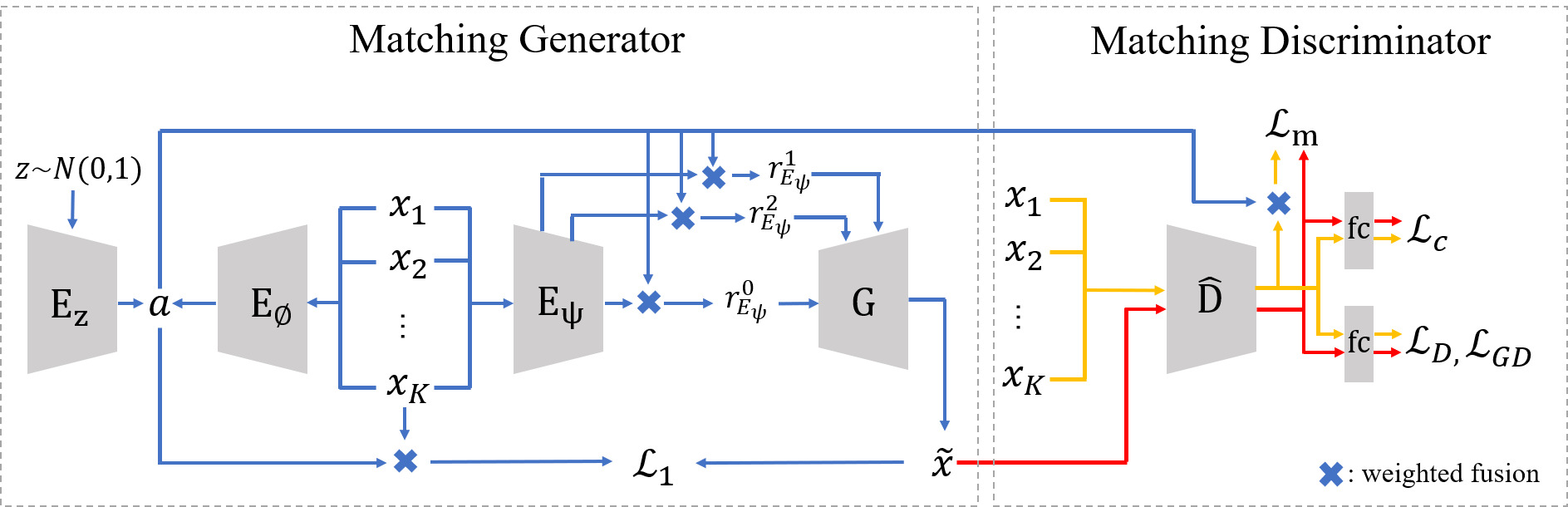}
\end{center}
\caption{The framework of our MatchingGAN which consists of a matching generator and a matching discriminator. $\tilde{\bm{x}}$ is generated based on the random vector $\bm{z}$ and $K$ conditional images $\{\bm{x}_i|_{i=1}^K\}$. Best viewed in color.}
\label{fig:framework} 
\end{figure*}

\section{Our MatchingGAN}
\label{sec:method}
\subsection{Overview}

 Our MatchingGAN aims to learn a mapping from a few conditional images $\mathcal{X}_S=\{\bm{x}_i|_{i=1}^K\}$ within one category to a new image $\tilde{\mathbf{x}}$ of this category, in which $K$ is the number of conditional images. Let $\mathcal{L}^{s}$ and $\mathcal{L}^{u}$ be the set of seen categories and unseen categories respectively, where $\mathcal{L}^{s} \cap \mathcal{L}^{u}=\emptyset$. In the training phase, MatchingGAN is trained on images from seen categories $\mathcal{L}^{s}$, without reaching images from unseen categories $\mathcal{L}^{u}$. The trained model owns the ability to fuse the information of conditional images to generate new images from the same category.

In the testing phase, given $K$ conditional images from an unseen category, the trained model could produce diverse and plausible images for this unseen category without any further fine-tuning. As illustrated in Fig.~\ref{fig:framework}, our MatchingGAN consists of a matching generator and a matching discriminator, which will be detailed next.

\subsection{Matching Generator}
In our matching generator, there are three encoders including $E_{z}$, $E_{\phi}$, and $E_{\psi}$ as well as a decoder $G$. Encoders $E_{z}$ and $E_{\phi}$ contribute to the matching procedure. Specifically, $E_{z}$ (\emph{resp.}, $E_{\phi}$) projects a random vector $\bm{z}$ sampled from unit Gaussian distribution $\mathcal{N}(\mathbf{0},\mathbf{1})$ (\emph{resp.}, conditional image $\bm{x}_i$) into a common matching space, leading to $E_{z}(\bm{z})$ (\emph{resp.}, $E_{\phi}(\bm{x}_i)$). In the matching space, we calculate the similarity score between random vector $\bm{z}$ and each conditional image $\bm{x}_i$ as
\begin{equation}
\begin{aligned}
\operatorname{a}(E_{z}(\bm{z}), E_{\phi}(\bm{x}_i))=\frac{{e}^{cos(E_{z}(\bm{z}),E_{\phi}(\bm{x}_i))}}{\sum_{i=1}^{K} {e}^{cos(E_{z}(\bm{z}), E_{\phi}(\bm{x}_i))}},
\end{aligned}
\end{equation}
in which $cos(\cdot,\cdot)$ is the cosine similarity. We calculate the normalized cosine similarity as similarity score, which are later used as interpolation coefficients to fuse the features of conditional images. Through this matching procedure, we tend to learn reasonable interpolation coefficients to determine how much information we should borrow from each conditional image.  Moreover, we expect that the matching procedure learned from seen categories can be adapted to unseen categories to 
generate images for unseen categories in the testing phase. Compared with simply using random interpolation coefficients without matching procedure, we verify that learning reasonable interpolation coefficients is more effective in our experiments (see Supplementary).

Encoder $E_{\psi}$ and decoder $G$ form an auto-encoder, which is used to generate new images based on the fused features of conditional images.
Our encoder $E_{\psi}$ and decoder $G$ are designed as UNet structure~\cite{ronneberger2015u}, in which the features of several blocks in encoder $E_{\psi}$ are connected to the output of corresponding blocks in the decoder $G$. Specifically, both $E_{\psi}$ and $G$ have four blocks, and we add skip connections between the final two blocks in $E_{\psi}$ and the first two blocks in $G$. We use $E_{\psi}^0(\bm{x}_i)$ to denote the bottleneck feature of conditional image $\bm{x}_i$, and use $E_{\psi}^j(\bm{x}_i)$ to denote the feature of the $j$-th encoder block with skip connection. After using similarity scores $\operatorname{a}(E_{z}(\bm{z}), E_{\phi}(\bm{x}_i))$ to interpolate the features $E_{\psi}^j(\bm{x}_i)$  of $K$ conditional images, we can have the fused features:
\begin{equation}
\begin{aligned}
\bm{r}_{E_{\psi}}^j =\sum_{i=1}^{K} \operatorname{a}(E_{z}(\bm{z}), E_{\phi}(\bm{x}_i)) E_{\psi}^j(\bm{x}_i),\,\, j=0,\ldots,L,  \label{a_g}
\end{aligned}
\end{equation}
in which $L$ is the number of skip connections.

Due to the skip connection between encoder $E_{\psi}$ and decoder $G$, we concatenate the fused feature $\bm{r}_{E_{\psi}}^j$ from $E_{\psi}^j$ and the output of its connected block in $G$ as the input to the next block in $G$. Then, the image generated by the decoder $G$ can be represented by
\begin{equation}
\begin{aligned}
\tilde{\bm{x}} = G(\bm{r}^0_{E_{\psi}}, \bm{r}^1_{E_{\psi}},\ldots,\bm{r}^L_{E_{\psi}}) . \label{generator}
\end{aligned}
\end{equation}

As the shallow (\emph{resp.}, deep) layer in the encoder integrates the low-level (\emph{resp.}, high-level) information, our generated images can fuse multi-level information of conditional images coherently based on the interpolated features in multiple layers. The effectiveness of fusing multi-level features is proved in our experiments (see Supplementary).

Considering that the generated image $\tilde{\bm{x}}$ fuses the information of $K$ conditional images $\bm{x}_i$ based on similarity scores $\operatorname{a}(E_{z}(\bm{z}), E_{\phi}(\bm{x}_i))$, we conjecture that $\tilde{\bm{x}}$ should appear more similar to the conditional image with higher similarity score. Thus, we employ the weighted reconstruction loss as follows,
\begin{equation}
\begin{aligned}
\mathcal{L}_1 = \sum_{i=1}^{K} \operatorname{a}(E_{z}(\bm{z}), E_{\phi}(\bm{x}_i)) || \bm{x}_i - \tilde{\bm{x}}||_1.
\end{aligned}
\end{equation}

The weighted reconstruction loss can make the network training more stable and enforce the generated images to contain the fused information of conditional images as expected.

\subsection{Matching Discriminator}
In our matching discriminator $D$, we treat $K$ conditional images $\bm{x}_i$ as real images while the generated images $\tilde{\bm{x}}$ as fake images. In detail, we calculate the average of scores $\mathrm{D}(\bm{x}_i)$ for $K$ conditional images $\bm{x}_i$ and score $\mathrm{D}(\tilde{\bm{x}})$ for the generated image $\tilde{\bm{x}}$. To stabilize adversarial learning, we adopt the hinge adversarial loss in~\cite{miyato2018cgans}. Concretely, the discriminator $\mathrm{D}$ tends to minimize the loss function $\mathcal{L}_D$ while the matching generator tends to minimize the loss function $\mathcal{L}_{GD}$:
\begin{eqnarray}
\!\!\!\!\!\!\!\!&&\mathcal{L}_D = \mathbb{E}_{\tilde{\bm{x}}} [\max (0,1+\mathrm{D}(\tilde{\bm{x}})]  +  \mathbb{E}_{\bm{x}_i}  [\max (0,1-\mathrm{D}({\bm{x}_i}))], \nonumber\\
\!\!\!\!\!\!\!\!&&\mathcal{L}_{GD} = - \mathbb{E}_{\tilde{\bm{x}}} [\mathrm{D}(\tilde{\bm{x}})].
\end{eqnarray}

Following ACGAN~\cite{odena2017conditional}, we construct classifier $C$ by replacing the last fully connected (fc) layer of the discriminator $D$ with another fc layer with $C^s$ outputs, in which $C^s$ is the number of seen categories. Then, we employ the cross-entropy classification loss to distinguish different categories:
\begin{equation}
\begin{aligned}
\mathcal{L}_{c} = -\log p(c(\bm{x})|\bm{x}),
\end{aligned}
\end{equation}
where $c(\bm{x})$ is the category of image $\bm{x}$. We train the discriminator by minimizing the classification  loss $\mathcal{L}^{D}_{c} = -\log p(c(\bm{x}_i)|\bm{x}_i)$ of conditional images $\mathcal{X}_S$. When updating the matching generator, we expect the generated image $\tilde{\bm{x}}$ to be classified as the same category of conditional images by minimizing the classification loss $\mathcal{L}^{G}_{c}=-\log p(c(\tilde{\bm{x}})|\tilde{\bm{x}})$.

In practice, the distributions of real and fake images may not overlap with each other, especially at the early stage of training process. Hence, the discriminator $D$ can separate them perfectly, which makes the training process of matching generator unstable. Considering the fact that the matching generator fuses the features of conditional images $\mathcal{X}_S$ according to the interpolation coefficients, to cooperate with the fusion strategy in the matching generator, we match the discriminative feature of $\tilde{\bm{x}}$ with the fused discriminative feature of $\mathcal{X}_S$, which is implemented by a feature matching loss. In detail, we remove the last fc layer of the discriminator $D$ and use the remaining feature extractor $\hat{D}$ to extract the discriminative features of generated images and conditional images. Thus, the feature matching loss can be written as
\begin{equation}
\begin{aligned}
\mathcal{L}_{m} = || \sum_{i=1}^{K} \operatorname{a}(E_{z}(\bm{z}), E_{\phi}(\bm{x}_i))  \hat{D}(\bm{x}_i)-  \hat{D}(\tilde{\bm{x}})||_1.
\end{aligned}
\end{equation}

\subsection{Optimization}
The total loss function our MatchingGAN can be written as 

\begin{equation}
\begin{aligned}
\mathcal{L} =  \mathcal{L}_D +  \mathcal{L}_{GD}+  \lambda_r \mathcal{L}_{1} + \mathcal{L}_{c} + \lambda_m \mathcal{L}_{m}.\label{optimization}
\end{aligned}
\end{equation}
During adversarial learning, matching generator and matching discriminator are optimized by different loss terms in an alternating manner. In particular, the matching discriminator $D$ is trained with $\mathcal{L}_D$ and $\mathcal{L}^{D}_c$, while the matching generator $E_{z}$, $E_{\phi}$, $E_{\psi}$, and $G$ are trained with $\mathcal{L}_{GD}$, $\mathcal{L}_{1}$, $\mathcal{L}^{G}_{c}$, and $\mathcal{L}_{m}$.

\section{Experiments}
In this section, we compare our MatchingGAN with existing methods by conducting generation and classification experiments on three datasets.
\label{sec:experiments}

\subsection{Datasets and Implementation Details}
Following DAGAN~\cite{antoniou2017data}, we conduct experiments on three datasets: Omniglot~\cite{Brenden2015One}, EMNIST~\cite{cohen2017emnist}, and VGGFace~\cite{cao2018vggface2}. We also follow the category split used in~\cite{antoniou2017data}. For Omniglot (\emph{resp.}, EMNIST, VGGFace), a total of $1623$ (\emph{resp.}, $48$, $2395$) categories are split into $1200$ (\emph{resp.}, $28$, $1802$) seen categories, $212$ (\emph{resp.}, $10$, $497$) validation seen categories, $211$ (\emph{resp.}, $10$ , $96$) unseen categories. Validation seen categories are used to monitor the training procedure, but not engaged in updating model parameters. For EMNIST and VGGFace, some categories have more than $100$ samples. Following \cite{antoniou2017data}, for these categories, we randomly choose $100$ images from each category to fit a low-data setting. 

We empirically set $\lambda_r = 0.1$ and $\lambda_m = 1$. The number of conditional images $K$ is set as $3$ considering the trade-off between effectiveness and efficiency.  We use Adam optimizer with learning rate 0.0001 and train our MatchingGAN for $200$ epochs.  The detailed architecture of matching generator and matching discriminator is provided in Supplementary.

\begin{figure}
\begin{center}
\includegraphics[scale=0.37]{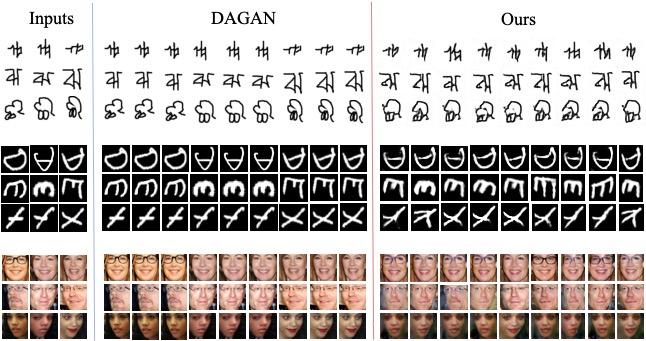}
\end{center}
\caption{Images generated by our MatchingGAN ($K=3$) and DAGAN on three datasets (from top to bottom: Omniglot, EMNIST, VGGFace). The conditional images are in the left three columns.}
\label{fig:visualization} 
\end{figure}

\setlength{\tabcolsep}{15pt}
\begin{table}[t]
  \caption{FID ($\downarrow$) and IS ($\uparrow$) of images generated by different methods on VGGFace dataset.} 
  \centering
  \begin{tabular}{|c|c|c|}
        \hline 
    	Methods & FID ($\downarrow$)  & IS ($\uparrow$) \cr
    \hline\hline
    FIGR~\cite{clouatre2019figr} & 154.21 & 5.19 \cr
    \hline
    GMN~\cite{bartunov2018few}& 201.12 & 6.38 \cr
    \hline
    DAGAN~\cite{antoniou2017data} & 121.43  & 4.12 \cr
    \hline\hline
    Ours &108.56  & 8.32 \cr
    \hline
  \end{tabular}
  \label{tab:performance_metric}
\end{table}

\setlength{\tabcolsep}{7pt}
\begin{table}[t]
  \caption{Accuracy(\%) of different methods on different datasets in low-data setting.}
  \centering
  
  \begin{tabular}{|c|c|c|c|c|}
	    	\hline
	    	\multirow{2}{*}{Method} & \multirow{2}{*}{Dataset} & \multicolumn{3}{c|}{Accuracy} \cr
			\cline{3-5}
			~& ~ & 5 & 10 &15 \cr

    \hline\hline
    Standard& Omniglot &66.22  & 81.87 &83.31 \cr
    \hline
    FIGR~\cite{clouatre2019figr}& Omniglot  & 69.23  & 83.12 & 84.89 \cr
    \hline
    GMN~\cite{bartunov2018few} &  Omniglot &  67.74 & 84.19 &  85.12 \cr
    \hline
    DAGAN~\cite{antoniou2017data}& Omniglot & 88.81  &89.32 & 95.38  \cr
    \hline
    Ours& Omniglot &89.03  &90.92 & 96.29  \cr
    \hline\hline
    Standard& EMNIST & 83.64 & 88.64 & 91.14 \cr
    \hline
    FIGR~\cite{clouatre2019figr}& EMNIST & 85.91  & 90.08  & 92.18 \cr
    \hline
    GMN~\cite{bartunov2018few}& EMNIST & 84.56  & 91.21 & 92.09 \cr
    \hline
    DAGAN~\cite{antoniou2017data}& EMNIST &87.45  & 94.18& 95.58 \cr
    \hline
    Ours& EMNIST & 91.75 & 95.91 &96.29  \cr
    \hline\hline
    Standard& VGGFace & 8.82 & 20.29 & 39.12\cr
    \hline
    FIGR~\cite{clouatre2019figr}& VGGFace & 6.12  &  18.84& 32.13 \cr
    \hline
    GMN~\cite{bartunov2018few}& VGGFace & 5.23  & 15.61 &35.48  \cr
    \hline
    DAGAN~\cite{antoniou2017data}& VGGFace &19.23 &35.12 & 44.36 \cr
    \hline
    Ours& VGGFace &21.12  &40.95 & 50.12 \cr
    \hline

  \end{tabular}
  \label{tab:performance_vallia_classifier}
\end{table}

\setlength{\tabcolsep}{2pt}
\begin{table}[t]
  \caption{Accuracy(\%) of different methods on different datasets in few-shot setting.} 
  \centering
  \begin{tabular}{|c|c|c|c|}
        \hline 
        Methods& Dataset & 5-way 5-shot &10-way 5-shot\cr
    \hline\hline
    MatchingNets~\cite{vinyals2016matching}&Omniglot  &98.70  &98.91 \cr
    \hline
    MAML~\cite{finn2017model}&Omniglot  &99.90   &99.13 \cr
    \hline
    RelationNets~\cite{sung2018learning}& Omniglot & 99.80 & 99.22 \cr
    \hline
    MTL~\cite{sun2019meta}& Omniglot  &99.85  &99.35 \cr
    \hline
    DN4~\cite{li2019revisiting}&Omniglot  &99.83  & 99.29\cr
    \hline
    Ours& Omniglot &99.93  &99.42 \cr
    \hline\hline

    MatchingNets~\cite{vinyals2016matching}&VGGFace  &60.01  & 48.67\cr
    \hline
    MAML~\cite{finn2017model}&VGGFace  & 61.09 & 47.89 \cr
    \hline
    RelationNets~\cite{sung2018learning}& VGGFace & 60.93 & 49.12 \cr
    \hline
    MTL~\cite{sun2019meta}& VGGFace  &63.67  &51.94 \cr
    \hline
    DN4~\cite{li2019revisiting}&VGGFace  &62.89  & 51.58 \cr
    \hline
    Ours& VGGFace & 65.12 &53.21 \cr
    \hline
    \end{tabular}
  \label{tab:performance_fewshot_classifier}
\end{table}

\subsection{Quantitative Evaluation of Generated Images} \label{sec:visualization}
We evaluate the quality of images generated by different methods on VGGFace dataset based on commonly used Inception Scores (IS)~\cite{xu2018empirical} and Fréchet Inception Distance (FID)~\cite{heusel2017gans}. The implementation details of IS and FID are described in Supplementary. 

For our MatchingGAN, we train the model based on seen categories. Then, we randomly select $K=3$ images from each unseen category, after which these conditional images and a random vector are fed into the trained model to generate a new image for this unseen category. We can repeat the above procedure to generate adequate images for each unseen category.
Similarly, we train GMN~\cite{bartunov2018few} and FIGR~\cite{clouatre2019figr} in $1$-way $3$-shot setting based on seen categories, and use trained model to generate  images for unseen categories. Distinctive from the above methods, DAGAN~\cite{antoniou2017data} is conditioned on a single image, but we can still  generate adequate images for unseen categories by using one conditional image each time.

We generate $128$ images for each unseen category using each method, based on which FID and IS are calculated. The results of different methods are reported in Table \ref{tab:performance_metric}, from which we observe that the images generated by our MatchingGAN achieve the highest IS and lowest FID, which demonstrates that our model could produce more diverse and realistic images compared with baseline methods.

For visualization comparison, we show some example images generated by our MatchingGAN on three datasets in Fig.~\ref{fig:visualization}. We also show the images generated by DAGAN for comparison, which is a competitive baseline as demonstrated in Table \ref{tab:performance_metric}. It can be seen that our method can produce more diverse images than DAGAN, because our method excels in fusing the information of more than one conditional image.
More visualization results can be found in Supplementary.

\subsection{Low-data Classification}
\label{sec:vallia}
To further evaluate the quality of generated images, we use generated images to help downstream classification tasks in low-data setting in this section and few-shot setting in Section \ref{sec:few-shot}. For low-data classification on unseen categories, we randomly select a few (\emph{e.g.}, $5$, $10$, $15$) training images per unseen category while the remaining images in each unseen category are test images. Note that we have training and testing phases for classification, which are different from the training and testing phases of our MatchingGAN.
We use ResNet$18$ ~\cite{he2016deep} pretrained on seen categories as  backbone network, train the classifier based on the training images of unseen categories, and finally predict the test images of unseen categories. This setting is referred to as ``standard" in Table~\ref{tab:performance_vallia_classifier}.

Then, we attempt to use generated images to augment the training set of unseen categories. For each few-shot generation method, we generate $512$ images for each unseen category based on the training set of unseen categories. Then, the ResNet$18$ classifier is trained on the augmented training set (original training set and generated images) and applied to the test set of unseen categories.
The results of different methods are listed in Table~\ref{tab:performance_vallia_classifier}. On Omniglot and EMNIST datasets,
all methods outperform ``standard", which demonstrates the benefit of augmented training set. On VGGFace dataset, our MatchingGAN and DAGAN~\cite{antoniou2017data} outperform ``standard",  while GMN and FIGR underperform ``standard". One possible explanation is that the images generated by GMN and FIGR on the more challenging VGGFace dataset are of low quality and mislead the training of ResNet$18$. It can also be seen that our proposed MatchingGAN achieves significant improvement over baseline methods, which corroborates the effectiveness of combining matching procedure with adversarial learning.

\subsection{Few-shot Classification}
\label{sec:few-shot}

Following the $N$-way $C$-shot setting in few-shot classification~\cite{vinyals2016matching,sung2018learning}, we create episodes and report the averaged accuracy over multiple episodes on each dataset. In each episode, we first randomly select $N$ unseen categories, and then randomly select $C$ images from each unseen category as training set and the remaining images are used as test set. Each episode is similar to the low-data setting in Section~\ref{sec:vallia}.
Again, we use ResNet$18$ as the classifier and generate $512$ images for each unseen category to augment the training set.

We compare our MatchingGAN with few-shot classification methods, including representative methods MatchingNets~\cite{vinyals2016matching}, RelationNets~\cite{sung2018learning}, MAML~\cite{finn2017model} as well as state-of-the-art methods MTL~\cite{sun2019meta}, DN4~\cite{li2019revisiting}. Note that the above few-shot classification methods do not generate images to augment the training set of unseen categories. Instead, we strictly follow their original training procedure based on seen categories and fine-tuning procedure based on the training set of unseen categories if necessary.

We conduct experiments in $5$-way/$10$-way $5$-shot setting on Omniglot and VGGFace datasets, and report the averaged results over $10$ episodes on each dataset. From Table~\ref{tab:performance_fewshot_classifier}, we can observe that our MatchingGAN achieves better results than few-shot classification methods, which shows the power of augmented images generated by our model.

\subsection{Ablation Studies}
We analyze the impact of hyper-parameters (\emph{i.e.}, $\lambda_r$, $\lambda_m$, $K$) in our method and investigate different design choices like skip connection. We also remove the matching procedure and use random interpolation coefficients to prove the necessity of matching procedure. Due to space limitation, we leave the detailed experimental results to Supplementary.

\section{Conclusion}
In this paper, we have proposed a novel few-shot generation method  MatchingGAN by combining matching procedure with adversarial learning. Comprehensive generation and classification experiments on three datasets have demonstrated that our MatchingGAN can generate more diverse and realistic images than existing methods.

\renewcommand\thesection{\Alph{section}}

\section{Details of Network Architecture}
\textbf{Matching Generator}
In our matching generator, $E_{\psi}$ and $G$ form an auto-encoder, which is a combination of UNet~\cite{ronneberger2015u} and ResNet~\cite{he2016deep}. Specifically, the auto-encoder has $8$ blocks ($4$ blocks for encoder and $4$ blocks for decoder), in which each block contains $4$ composite layers (leaky ReLU and batch normalization followed by one downscaling or upscaling layer). Downscaling layers (in blocks $1-4$) are convolutional layers with stride $2$ followed by leaky ReLU, batch normalization, and dropout. Upscaling layers (in blocks $5-8$) are stride $1/2$ replicators followed by a convolutional layer, leaky ReLU, batch normalization, and dropout. The first $2$ blocks of encoder and the last $2$ blocks of decoder have $64$ convolutional filters, while the last $2$ blocks of the encoder and the first $2$ blocks of the decoder have $128$ convolutional filters. Skip connections are added between the last two blocks of encoder and the first two blocks of decoder.

For matching procedure, $E_{\phi}$ has the same structure and shared model parameters with $E_{\psi}$.  $E_z$ only has a fully connected (fc) layer with $d$ outputs, where $d$ is the same dimension as the output of encoder $E_{\phi}$.

\noindent \textbf{Matching Discriminator}
The network structure of our disciminator is similar to that in~\cite{liu2019few}. The discriminator consists of one convolutional layer followed by five blocks with increasing numbers of channels. The structure of each block is as follows: ResBlk-$k$ $\rightarrow$ ResBlk-$k$ $\rightarrow$ AvePool$2$x$2$, where ResBlk-$k$ is a ReLU first residual block~\cite{mescheder2018training} with the number of channels $k$ set as $64$, $128$, $256$, $512$, $1024$ in five blocks. We use one fully connected (fc) layer with $1$ output following AvePool layer to obtain the discriminator score. The classifier shares the feature extractor with discriminator and only replaces the last fc layer with another fc layer with $C^s$ outputs with $C^s$ being the number of seen categories. To obtain the features for feature matching loss, we remove the last fc layer from discriminator to extract the discriminative features of conditional image $\mathcal{X}_S$ and generated image $\tilde{\bm{x}}$.

\section{Details of Performance Metrics}
\textbf{Inception Scores}
We use the Inception Score (IS)~\cite{xu2018empirical}, which is widely used for evaluating the quality of generated images. Let $p(y|\tilde{\bm{x}})$ be the posterior distribution of generated image $\tilde{\bm{x}}$ over unseen categories. The inception score is given by:
\begin{equation}
\begin{aligned}
\mathrm{IS}=\exp \left(\mathrm{E}_{\tilde{\bm{x}} \sim \mathbb{p}(\tilde{\bm{x}})}[\mathrm{KL}(\mathbb{p}(y | \tilde{\bm{x}}) | \mathbb{p}(y)]\right)
\end{aligned}
\end{equation}
where $\mathbb{p}(y)=\int_{\tilde{\bm{x}}} \mathbb{p}(y | \tilde{\bm{x}}) d \tilde{\bm{x}}$. It is argued that the inception score is positively correlated with visual quality of generated images.

\noindent \textbf{Fréchet Inception Distance}
The Frechet Inception Distance (FID)~\cite{heusel2017gans} is designed for measuring similarities between two sets of images. We remove the last average pooling layer of the ImageNet-pretrained Inception-V3~\cite{szegedy2016rethinking} model as the feature extractor. Then, we compute FID between the generated images and the real images from the unseen categories. 

\section{More Example Images Generated by Our MatchingGAN}
\begin{figure*}
\begin{center}
\includegraphics[scale=0.8]{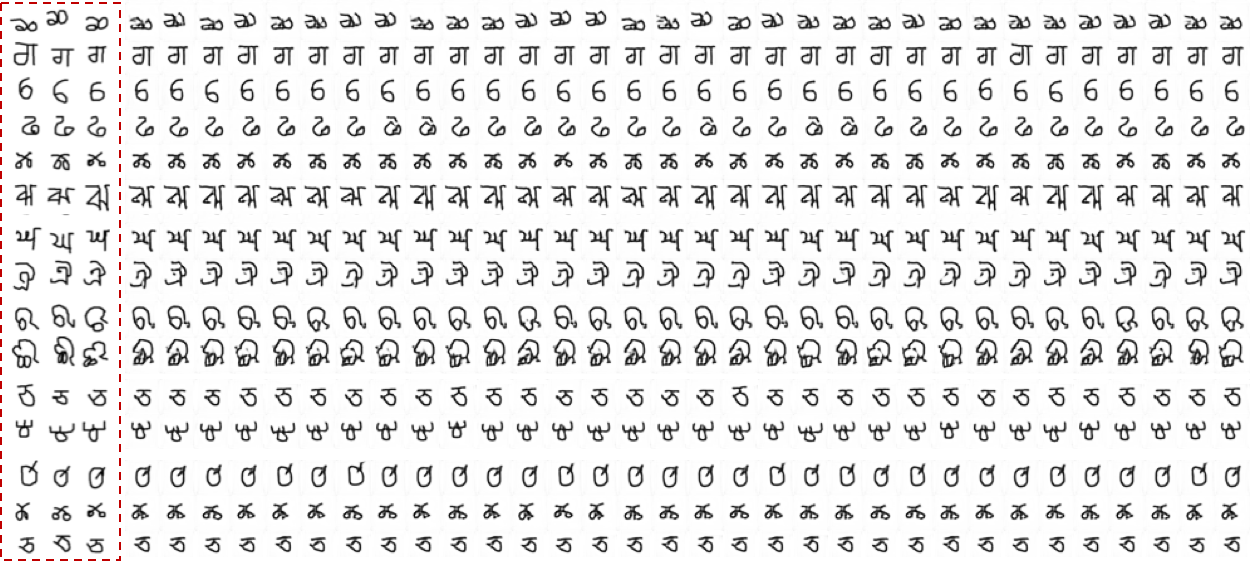}
\end{center}
\caption{Images generated from trained MatchingGAN with $K=3$ on Omniglot Dataset. The real conditional images are the left three columns.}
\label{fig:omni} 
\end{figure*}

\begin{figure*}
\begin{center}
\includegraphics[scale=0.8]{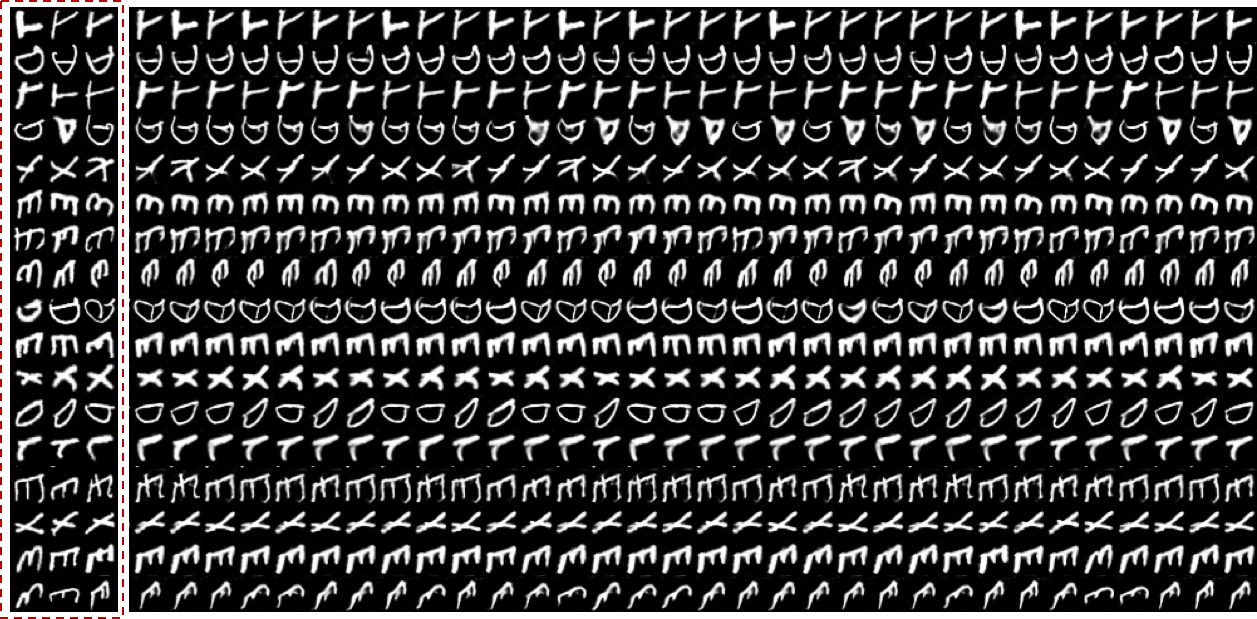}
\end{center}
\caption{Images generated from trained MatchingGAN with $K=3$ on EMNIST Dataset. The real conditional images are the left three columns.}
\label{fig:emni} 
\end{figure*}

\begin{figure*}
\begin{center}
\includegraphics[scale=0.8]{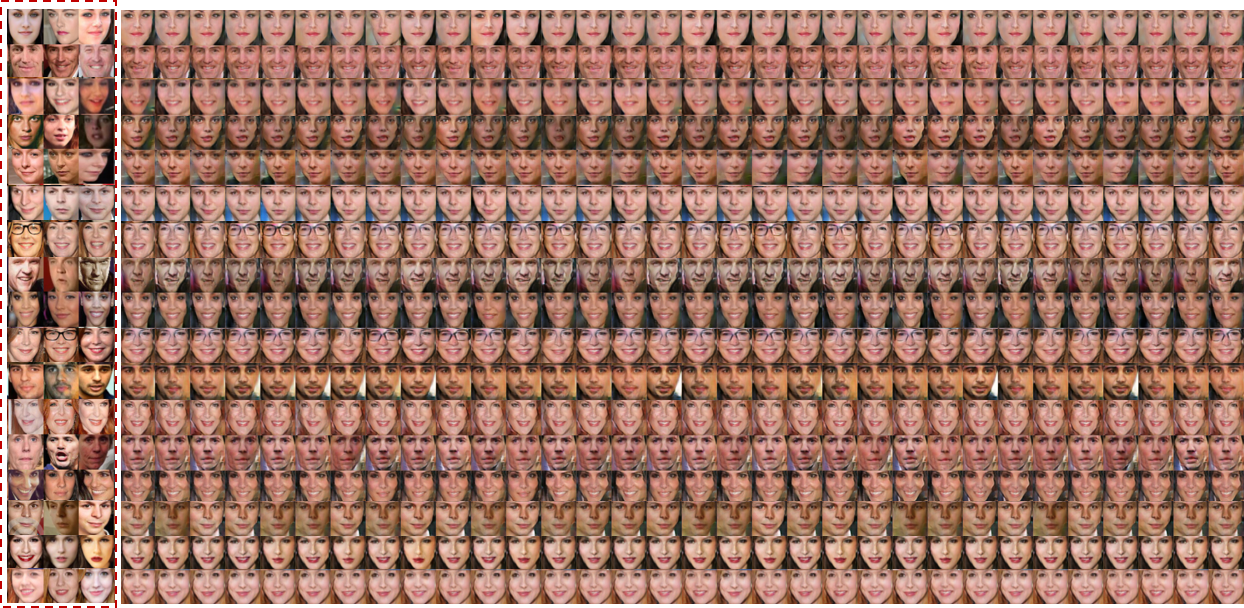}
\end{center}
\caption{Images generated from trained MatchingGAN with $K=3$ on VGGFace Dataset. The real conditional images are the left three columns.}
\label{fig:face} 
\end{figure*}

\begin{figure*}
\begin{center}
\includegraphics[scale=0.8]{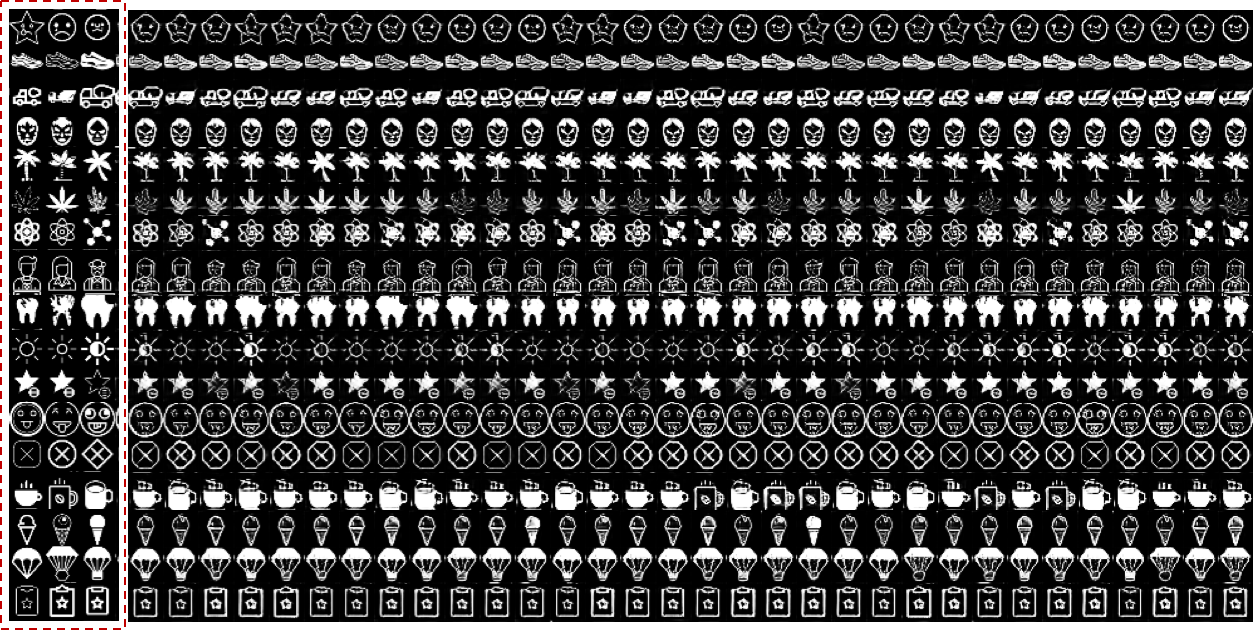}
\end{center}
\caption{Images generated from trained MatchingGAN with $K=3$ on FIGR Dataset. The real conditional images are the left three columns.}
\label{fig:figr} 
\end{figure*}

We show more example images generated by our MatchingGAN ($K=3$) on Omniglot, EMNIST, and VGGFace datasets in Fig.~\ref{fig:omni}, Fig.~\ref{fig:emni}, and Fig.~\ref{fig:face} respectively. Besides, we additionally conduct experiments on FIGR~\cite{clouatre2019figr} dataset, which is not used in our main paper. The generated images on FIGR dataset are shown in Fig.~\ref{fig:figr}. On all four datasets, our MatchingGAN can generate diverse and plausible images based on a few conditional images.

\setlength{\tabcolsep}{8pt}
\begin{table}[t]
  \caption{Accuracy(\%) of low-data (10-sample) classification augmented by our MatchingGAN with different $K_1$ and $K_2$ on EMNIST dataset.} 
  \centering
  \begin{tabular}{|c|c|c|c|c|}
  \hline
		~ & $K_1=3$ & $K_1=5$&$K_1=7$ & $K_1=9$  \cr
    	
    \hline
     $K_2=3$&95.91  & 95.72  & 94.89 & 93.96  \cr
    \hline
     $K_2=5$& 94.01 &  96.11 & 95.79 & 95.08  \cr
    \hline
      $K_2=7$&92.89  &  93.89 & 96.92 & 96.16  \cr
    \hline
      $K_2=9$&88.42  & 90.12  &  92.21& 97.11  \cr
    \hline    
  \end{tabular}
  \label{tab:number_effect}
\end{table}

\section{Ablation studies}
\textbf{The number of conditional images} To analyze the impact of the number of conditional images, we train MatchingGAN with $K_1$ conditional images based on seen categories, and generate new images for unseen categories with $K_2$ conditional images. By default, we set $K=K_1=K_2=3$ in our experiments.
We evaluate the quality of generated images using different $K_1$ and $K_2$ in low-data (\emph{i.e.}, $10$-sample) classification, which is the same as Section $4.3$ in the main paper.
By taking EMNIST dataset as an example, we report the results in Table~\ref{tab:number_effect} by varying $K_1$ and $K_2$ in the range of $[3, 5, 7, 9]$. From Table~\ref{tab:number_effect}, we can observe that when $K_2=K_1$, our MatchingGAN can generally achieve good performance and the performance increases as $K$ increases. Besides, we observe that given a fixed $K_2$, when $K_1>K_2$, the performance is not degraded a lot compared with $K_2=K_1$. However, given a fixed $K_2$, when $K_1<K_2$, the performance is significantly degraded. We conjecture that if our MatchingGAN is trained with $K_1$ conditional images, it cannot generalize well to fuse the information of more conditional images ($K_2>K_1$) in the testing phase.

\noindent\textbf{Hyper-parameter analysis} In our MatchingGAN, we add a hyper-parameter $\lambda_r$ before the weighted reconstruction loss $\mathcal{L}_{1}$ and a hyper-parameter $\lambda_m$ before the feature matching loss $\mathcal{L}_{m}$. We investigate the impact of hyper-parameters $\lambda_r$ and $\lambda_m$ on VGGFace dataset, by varying $\lambda_r$ (\emph{resp.}, $\lambda_m$) in the range of $[0.01, 0.1, 1, 10]$ (\emph{resp.}, $[0, 1]$). We evaluate the quality of generated images from two perspectives. On one hand, we compute the Inception Score (IS) and Fréchet Inception Distance (FID) of generated images as done in Section $4.2$ in the main paper. On the other hand, we report the accuracy of low-data ($10$-sample) classification augmented with generated images as described in Section $4.3$ in the main paper. The results are reported in Table~\ref{tab:network_design}, which shows that larger $\lambda_r$ leads to lower FID and higher IS at the cost of classification performance. $\lambda_r=0.1$ achieves a good trade-off, so we use $\lambda_r=0.1$ as default value in our experiments. Another observation is that after removing the feature matching loss by setting $\lambda_m=0$, IS, FID, and classification accuracy become significantly worse, which indicates the benefit of feature matching loss. 

\noindent\textbf{Random interpolation coefficient} In our MatchingGAN, we employ matching procedure to learn reasonable interpolation coefficients, which are used to fuse the features of conditional images. A naive alternative to the matching procedure is to sample normalized random vector $[a_1, a_2,\ldots,a_K]$ with $a_i>=0$ and $\sum_i a_i=1$ from uniform distribution as interpolation coefficients. In this way, we can use random interpolation coefficients in both training and testing phase, so that the matching procedure could be discarded. In particular, $E_{z}$ and $E_{\phi}$ will not participate in generator training. When using the trained generator to generate new images for unseen categories, we can also randomly sample interpolation coefficients without relying on $E_{z}$ and $E_{\phi}$.

By taking VGGFace dataset as an example, we compare MatchingGAN with matching procedure with the one without matching procedure (use random interpolation coefficients) based on three evaluation metrics (\emph{i.e.}, FID, IS, and low-data classification accuracy). The results are listed in Table~\ref{tab:network_design}, which demonstrate that matching procedure is capable of learning more reasonable interpolation coefficients than random interpolation coefficients, leading to better generated images.

\noindent\textbf{Network design choices} To investigate surrogate choices of network design, we again take VGGFace dataset as an example and utilize three evaluation metrics (\emph{i.e.}, FID, IS, and low-data classification accuracy) for comparison. In our MatchingGAN, the encoder $E_{\phi}$ has the same network structure as $E_{\psi}$ with shared model parameters. Alternatively, we can learn different model parameters for $E_{\phi}$ and $E_{\psi}$ separately. Table ~\ref{tab:network_design} records the results of MatchingGAN in these two different cases. We observe that although introducing more model parameters, learning two encoders separately does not notably improve the performance of MatchingGAN. 

Besides, in our matching generator, the number of skip connections between encoder $E_{\psi}$ and decoder $G$ also affects the quality of fused features and generated images. We utilize two connection blocks by default in our experiments. Here, we further explore the effect of using different numbers of skip connections. We report the results using $1$, $2$, $3$ skip connections in Table~\ref{tab:network_design}. For $1$ skip connection, we only keep the skip connection between the last block in encoder $E_{\psi}$ and the first block in decoder $G$. For $3$ skip connections, we add another connection between the second block in encoder $E_{\psi}$ and the third block in decoder $G$. According to Table~\ref{tab:network_design}, we can see that using more skip connections could improve the realism of generated images (lower FID and higher IS). Another observation is that $3$ skip connections compromise the low-data classification performance, because the generated images become closer to conditional images and lacking of diversity based on our experimental observation. We conjecture that it would be better to fuse multi-level information with an appropriate number of skip connections, so we opt for two skip connections in our matching generator.

 \begin{table}[t]
  \caption{Analyses of hyper-parameters and different network design choices on VGGFace dataset.} 
  \centering
  \begin{tabular}{|c|c|c|c|}  
    \hline
     setting& accuracy  & FID ($\downarrow$)  & IS ($\uparrow$)  \cr
     \hline
    \hline
     $\lambda_r=0.01$ &35.62  & 112.16   & 7.89   \cr
    \hline
     $\lambda_r=0.1$ &40.95  &  108.56 & 8.32    \cr
    \hline
     $\lambda_r=1$ &33.89  &  107.16   & 9.17     \cr
     \hline
    $\lambda_r=10$ &30.12  &  106.12 & 11.04 \cr
    \hline   
    \hline
    $\lambda_m=1$ &40.95  &  108.56 & 8.32 \cr  
    \hline 
    $\lambda_m=0$ &28.98  &  111.4 & 7.56 \cr
    \hline
    \hline
    matching coefficient&40.95  &  108.56 & 8.32 \cr  
    \hline
    random coefficient&38.12  &  110.98 & 7.92 \cr
    \hline
    \hline
    shared encoder&40.95  &  108.56 & 8.32 \cr  
    \hline
    different encoder&40.98  &  107.98 & 8.56 \cr
    \hline   
    \hline
    1 connection &38.67  &  113.21 & 7.09 \cr
    \hline
    2 connection &40.95  &  108.56 & 8.32  \cr
    \hline
    3 connection &34.12  &  106.12 & 9.14 \cr
    \hline
  \end{tabular}
  \label{tab:network_design}
\end{table}

\begin{small}
\bibliographystyle{IEEEbib}
\bibliography{icme2020template}

\begin{thebibliography}{10}

\bibitem{bao2017cvae}
Jianmin Bao, Dong Chen, Fang Wen, Houqiang Li, and Gang Hua,
\newblock ``{CVAE-GAN}: fine-grained image generation through asymmetric
  training,''
\newblock in {\em ICCV}, 2017.

\bibitem{goodfellow2014generative}
Ian Goodfellow, Jean Pouget-Abadie, Mehdi Mirza, Bing Xu, David Warde-Farley,
  Sherjil Ozair, Aaron Courville, and Yoshua Bengio,
\newblock ``Generative adversarial nets,''
\newblock in {\em NeurIPS}, 2014.

\bibitem{arjovsky2017wasserstein}
Martin Arjovsky, Soumith Chintala, and L{\'e}on Bottou,
\newblock ``Wasserstein generative adversarial networks,''
\newblock in {\em ICML}, 2017.

\bibitem{clouatre2019figr}
Louis Clou{\^a}tre and Marc Demers,
\newblock ``Figr: Few-shot image generation with reptile,''
\newblock {\em arXiv preprint arXiv:1901.02199}, 2019.

\bibitem{bartunov2018few}
Sergey Bartunov and Dmitry Vetrov,
\newblock ``Few-shot generative modelling with generative matching networks,''
\newblock in {\em ICAIS}, 2018.

\bibitem{schwartz2018delta}
Eli Schwartz, Leonid Karlinsky, Joseph Shtok, Sivan Harary, Mattias Marder,
  Abhishek Kumar, Rogerio Feris, Raja Giryes, and Alex Bronstein,
\newblock ``Delta-encoder: an effective sample synthesis method for few-shot
  object recognition,''
\newblock in {\em NeurIPS}, 2018.

\bibitem{antoniou2017data}
Antreas Antoniou, Amos Storkey, and Harrison Edwards,
\newblock ``Data augmentation generative adversarial networks,''
\newblock {\em arXiv preprint arXiv:1711.04340}, 2017.

\bibitem{nichol2018first}
Alex Nichol, Joshua Achiam, and John Schulman,
\newblock ``On first-order meta-learning algorithms,''
\newblock {\em arXiv preprint arXiv:1803.02999}, 2018.

\bibitem{vinyals2016matching}
Oriol Vinyals, Charles Blundell, Timothy Lillicrap, Daan Wierstra, et~al.,
\newblock ``Matching networks for one shot learning,''
\newblock in {\em NeurIPS}, 2016.

\bibitem{miyato2018spectral}
Takeru Miyato, Toshiki Kataoka, Masanori Koyama, and Yuichi Yoshida,
\newblock ``Spectral normalization for generative adversarial networks,''
\newblock {\em arXiv preprint arXiv:1802.05957}, 2018.

\bibitem{miyato2018cgans}
Takeru Miyato and Masanori Koyama,
\newblock ``c{GAN}s with projection discriminator,''
\newblock {\em arXiv preprint arXiv:1802.05637}, 2018.

\bibitem{lu2018image}
Yongyi Lu, Shangzhe Wu, Yu-Wing Tai, and Chi-Keung Tang,
\newblock ``Image generation from sketch constraint using contextual {GAN},''
\newblock in {\em ECCV}, 2018.

\bibitem{liu2019few}
Ming-Yu Liu, Xun Huang, Arun Mallya, Tero Karras, Timo Aila, Jaakko Lehtinen,
  and Jan Kautz,
\newblock ``Few-shot unsupervised image-to-image translation,''
\newblock {\em arXiv preprint arXiv:1905.01723}, 2019.

\bibitem{sung2018learning}
Flood Sung, Yongxin Yang, Li~Zhang, Tao Xiang, Philip~HS Torr, and Timothy~M
  Hospedales,
\newblock ``Learning to compare: Relation network for few-shot learning,''
\newblock in {\em CVPR}, 2018.

\bibitem{zheng2019relation}
Wenbo Zheng, Lan Yan, Chao Gou, Wenwen Zhang, and Fei-Yue Wang,
\newblock ``A relation network embedded with prior features for few-shot
  caricature recognition,''
\newblock in {\em ICME}, 2019.

\bibitem{santoro2016one}
Adam Santoro, Sergey Bartunov, Matthew Botvinick, Daan Wierstra, and Timothy
  Lillicrap,
\newblock ``One-shot learning with memory-augmented neural networks,''
\newblock {\em arXiv preprint arXiv:1605.06065}, 2016.

\bibitem{finn2017model}
Chelsea Finn, Pieter Abbeel, and Sergey Levine,
\newblock ``Model-agnostic meta-learning for fast adaptation of deep
  networks,''
\newblock in {\em ICML}, 2017.

\bibitem{du2019low}
Xuefeng Du, Dexing Zhong, and Pengna Li,
\newblock ``Low-shot palmprint recognition based on meta-siamese network,''
\newblock in {\em ICME}, 2019.

\bibitem{Krizhevsky2012ImageNet}
Alex Krizhevsky, I.~Sutskever, and G.~Hinton,
\newblock ``Imagenet classification with deep convolutional neural networks,''
\newblock {\em NeurIPS}, 2012.

\bibitem{hariharan2017low}
Bharath Hariharan and Ross Girshick,
\newblock ``Low-shot visual recognition by shrinking and hallucinating
  features,''
\newblock in {\em ICCV}, 2017.

\bibitem{wang2018low}
Yu-Xiong Wang, Ross Girshick, Martial Hebert, and Bharath Hariharan,
\newblock ``Low-shot learning from imaginary data,''
\newblock in {\em CVPR}, 2018.

\bibitem{ronneberger2015u}
Olaf Ronneberger, Philipp Fischer, and Thomas Brox,
\newblock ``U-net: Convolutional networks for biomedical image segmentation,''
\newblock in {\em MICCAI}. Springer, 2015.

\bibitem{odena2017conditional}
Augustus Odena, Christopher Olah, and Jonathon Shlens,
\newblock ``Conditional image synthesis with auxiliary classifier {GAN}s,''
\newblock in {\em ICML}, 2017.

\bibitem{Brenden2015One}
Brenden~M. Lake, Ruslan Salakhutdinov, and Joshua~B. Tenenbaum,
\newblock ``One-shot learning by inverting a compositional causal process,''
\newblock {\em NeurIPS}, 2015.

\bibitem{cohen2017emnist}
Gregory Cohen, Saeed Afshar, Jonathan Tapson, and Andr{\'e} van Schaik,
\newblock ``{EMNIST}: an extension of {MNIST} to handwritten letters,''
\newblock {\em arXiv preprint arXiv:1702.05373}, 2017.

\bibitem{cao2018vggface2}
Qiong Cao, Li~Shen, Weidi Xie, Omkar~M Parkhi, and Andrew Zisserman,
\newblock ``Vggface2: A dataset for recognising faces across pose and age,''
\newblock in {\em FG}, 2018.

\bibitem{sun2019meta}
Qianru Sun, Yaoyao Liu, Tat-Seng Chua, and Bernt Schiele,
\newblock ``Meta-transfer learning for few-shot learning,''
\newblock in {\em CVPR}, 2019.

\bibitem{li2019revisiting}
Wenbin Li, Lei Wang, Jinglin Xu, Jing Huo, Yang Gao, and Jiebo Luo,
\newblock ``Revisiting local descriptor based image-to-class measure for
  few-shot learning,''
\newblock in {\em CVPR}, 2019.

\bibitem{xu2018empirical}
Qiantong Xu, Gao Huang, Yang Yuan, Chuan Guo, Yu~Sun, Felix Wu, and Kilian
  Weinberger,
\newblock ``An empirical study on evaluation metrics of generative adversarial
  networks,''
\newblock {\em arXiv preprint arXiv:1806.07755}, 2018.

\bibitem{heusel2017gans}
Martin Heusel, Hubert Ramsauer, Thomas Unterthiner, Bernhard Nessler, and Sepp
  Hochreiter,
\newblock ``{GAN}s trained by a two time-scale update rule converge to a local
  nash equilibrium,''
\newblock in {\em NeurIPS}, 2017.

\bibitem{he2016deep}
Kaiming He, Xiangyu Zhang, Shaoqing Ren, and Jian Sun,
\newblock ``Deep residual learning for image recognition,''
\newblock in {\em CVPR}, 2016.

\bibitem{mescheder2018training}
Lars Mescheder, Andreas Geiger, and Sebastian Nowozin,
\newblock ``Which training methods for {GAN}s do actually converge?,''
\newblock {\em arXiv preprint arXiv:1801.04406}, 2018.

\bibitem{szegedy2016rethinking}
Christian Szegedy, Vincent Vanhoucke, Sergey Ioffe, Jon Shlens, and Zbigniew
  Wojna,
\newblock ``Rethinking the inception architecture for computer vision,''
\newblock in {\em CVPR}, 2016.

\end{thebibliography}
\end{small}

\end{document}